\documentclass{article}
\usepackage{ijcai25}

\usepackage{times}
\usepackage{soul}
\usepackage{url}
\usepackage[hidelinks]{hyperref}
\usepackage[utf8]{inputenc}
\usepackage[small]{caption}
\usepackage{graphicx}
\usepackage{amsmath}
\usepackage{amsthm}
\usepackage{booktabs}
\usepackage{algorithm}
\usepackage{algorithmic}
\usepackage[switch]{lineno}

\usepackage{graphicx}
\usepackage{amsmath}
\usepackage{hyperref}
\usepackage{appendix}
\usepackage{multirow}
\usepackage{tabularx}
\usepackage{makecell}
\usepackage{algorithm}
\usepackage{algorithmic}
\usepackage{xcolor}
\usepackage{marvosym}
\usepackage{amssymb}

\title{FGeo-HyperGNet: Geometric Problem Solving Integrating FormalGeo Symbolic System and Hypergraph Neural Network}

\author{
Xiaokai Zhang$^1$ \and
Yang Li$^1$ \and
Na Zhu$^2$ \and
Cheng Qin$^2$ \and
Zhenbing Zeng$^3$ \And
Tuo Leng$^{1,2,}$\thanks{Corresponding author.} \\
\affiliations
$^1$School of Computer Engineering and Science, Shanghai University \\
$^2$School of Future Technology, Shanghai University \\
$^3$College of Sciences, Shanghai University \\
\emails
\{xiaokaizhang, leeyoung628, nazhu, qincheng, zbzeng, tleng\}@shu.edu.cn,
}

\begin{document}

\maketitle

\begin{abstract}
Geometric problem solving has always been a long-standing challenge in the fields of mathematical reasoning and artificial intelligence. We built a neural-symbolic system, called FGeo-HyperGNet, to automatically perform human-like geometric problem solving. The symbolic component is a formal system built on FormalGeo, which can automatically perform geometric relational reasoning and algebraic calculations and organize the solution into a hypergraph with conditions as hypernodes and theorems as hyperedges. The neural component, called HyperGNet, is a hypergraph neural network based on the attention mechanism, including an encoder to encode the structural and semantic information of the hypergraph and a theorem predictor to provide guidance in solving problems. The neural component predicts theorems according to the hypergraph, and the symbolic component applies theorems and updates the hypergraph, thus forming a predict-apply cycle to ultimately achieve readable and traceable automatic solving of geometric problems. Experiments demonstrate the effectiveness of this neural-symbolic architecture. We achieved state-of-the-art results with a TPA of 93.50\% and a PSSR of 88.36\% on the FormalGeo7K dataset. The code is available at \url{https://github.com/BitSecret/HyperGNet}.
\end{abstract}


\section{Introduction}

Geometry problem solving (GPS) has always been a long-standing challenge \cite{littman2022gathering,gowers2023artificial} in the fields of mathematical reasoning and artificial intelligence, due to the cross-modal forms of knowledge and the symbolic-numerical hybrid reasoning process. GPS can be described as: Given a geometric problem description (original images and texts or formalized), the solver needs to implement stepwise reasoning leading to the final answer.

\begin{figure*}
\centering
\includegraphics[width=\textwidth]{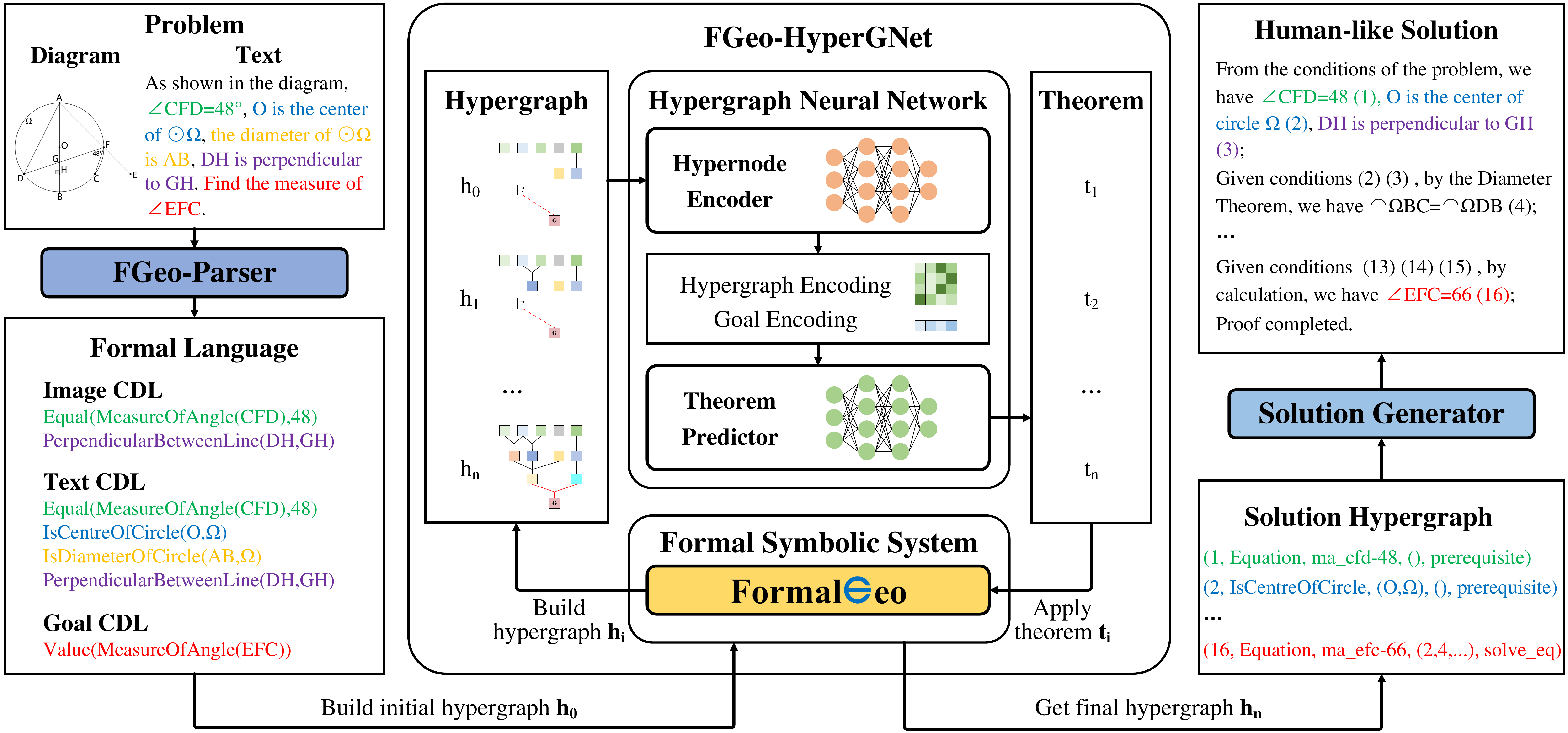}
\caption{The overall architecture of our proposed neural-symbolic system.}
\label{fig:architecture}
\end{figure*}

Traditional GPS methods can generally be divided into three categories. The first category is the synthesis methods, such as backward search method \cite{gelernter1959realization}, forward chaining method \cite{nevins1975plane} and deductive database method \cite{chou2000deductive}; the second category is the algebraic methods, such as Wu's method \cite{wu1978decision} and Gr{\"o}bner bases method \cite{buchberger1988applications}; the third category is the point elimination methods based on geometric invariants \cite{zhang1995automated,chou1995automated}.

Artificial intelligence technology has provided new perspectives for GPS \cite{seo2015solving,sachan2017learning,gan2019automatically}. In particular, with the rapid development of deep learning and the application of large language models, a series of neural-symbolic methods have been proposed. These methods can generally be divided into two categories: Deductive Database methods (DD methods) \cite{lu2021inter,peng2023geodrl,trinh2024solving,wu2024egps} and Program Sequence Generation methods (PSG methods) \cite{chen2021geoqa,zhang2023multi,xiao2024learning}. DD methods parse the problem images and texts into a unified formal language description, and then apply a predefined set of theorems to solve the problems. These approaches require the establishment of a formal system and the problem-solving process has mathematical rigor and good readability. PSG methods view GPS as a sequence generation task with multimodal input. These methods learn from annotated examples to map geometric problem descriptions into executable programs. After program sequences are generated, the executor computes them step by step and obtain the problem answer.

However, existing research has notable limitations. Most recent advances in GPS focus on exploring new methods and models \cite{xiao2023deep}, yet they overlook the investigation of geometric formal systems. The theorems and predicates of these formal systems are implemented using programming languages, and the definition of new predicates and theorems necessitates modifications to the solver's code. This characteristic significantly hampers the scalability of formal systems. Most Existing DD methods are non-traceable, and the redundant theorems applied during heuristic searches cannot be eliminated. Most Existing PSG methods, on the other hand, fail to yield a human-like problem-solving process, suffer from low readability, and cannot guarantee the correctness of results. The process of solving geometric problems encompasses both numerical computation and relational reasoning, with existing research predominantly focused on numerical problem-solving objectives, struggling to integrate computation and reasoning within a unified framework \cite{chen2022unigeo}. This imperfection in the existing formal systems severely restricts the types and complexity of GPS.

In addition, existing work predominantly focuses on the unified cross-modal integration of geometric text and image \cite{zhang2023multi,zhang2023lans} with limited attention to the embedding of geometric formal languages. AlphaGeometry \cite{trinh2024solving} can solve IMO-level geometry problems, but it treats the solving process as a text sequence and models GPS as a text generation task, ignoring structural relationships between conditions. Neglecting of graph structure information in formal language results in poor theorem prediction \cite{guo2022graph}. S2G \cite{tsai2021sequence} maps the problem-solving process onto an expression tree, implicitly incorporating process information, yet it does not reflect a human-like problem-solving approach. GeoDRL \cite{peng2023geodrl} organizes geometric conditions into a Geometric Logic Graph (GLG), but the GLG lacks information about the problem-solving process thus fails to model the interrelations among theorems. Formal language, distinct from natural language, adheres to stringent syntactic forms. Its symbols bear specific meanings and inappropriate tokenization can obliterate the inherent meaning of statements \cite{ning2023symbolic}. Moreover, due to the unique structure of formal languages, they are represented as three-dimensional real number matrices, which cannot be processed using common network architectures. There is an urgent need for research into the embedding and encoding of formal languages.

We propose a neural-symbolic architecture, named FGeo-HyperGNet, to address these issues, as illustrated in Figure~\ref{fig:architecture}. \textbf{The neural component} is a hypergraph neural network based on the attention mechanism, consisting of a hypernode encoder and a theorem predictor. The hypernode encoder embeds the semantic information of hypernodes and the neighboring edge information into fixed-length real-number vectors, serving as the feature representations of the nodes. The theorem predictor adopts an encoder-decoder architecture. It first extracts and fuses the node features obtained by the hypernode encoder to generate a hypergraph encoding. Subsequently, it uses a task-specific decoder that receives the hypergraph encoding and the goal encoding to predict the theorems required for solving geometric problems. We first use a self-supervised method to pretrain the hypernode encoder, making it to retain as much semantic information of formalized statements as possible during the encoding stage. Then the encoder is build as part of the theorem predictor for end-to-end training. \textbf{The symbolic component} is a symbolic formal system built on FormalGeo \cite{zhang2024formal}, which can construct the process of GPS as a directed hypergraph with conditions as hypernodes and theorems as hyperedges. This symbolic system can validate and apply the theorems predicted by the neural component, perform geometric relational reasoning and algebraic equation solving, and update the state of the hypergraph. 

The neural component predicts theorems according to the hypergraph, and the symbolic component applies theorems and updates the hypergraph, thus forming a predict-apply cycle (PAC) to ultimately achieve readable, traceable and verifiable automatic solving of geometric problems. Additionally, we utilize FGeo-Parser \cite{zhu2024fgeoparser} to convert geometric problem images and text into formalized language. Benefiting from the structured representation of the problem-solving process, we define a rule-based Solution Generator to derive a human-like solution. To the best of our knowledge, we are the first to construct a geometric problem-solving system that takes raw geometric problem images and text as input, produces a human-like solution as output, and ensures the complete correctness of the solution. Our work presents a neuro-symbolic framework for GPS, while demonstrating that hypergraph-structured data can enhance the capabilities of such methods. 

Our contributions are summarized as follows:

1. We introduce HyperGNet, an attention-based hypergraph feature embedding and extraction network. Unlike message-passing models, HyperGNet prioritizes the global relationships and representations of hypernodes, which are crucial for addressing the GPS task.

2. We present FGeo-HyperGNet, a neural-symbolic architecture designed for GPS. The neural component predicts theorems required to solve geometric problems based on the hypergraph, while the symbolic component conducts rigorous geometric relational reasoning and algebraic equation solving to ensure the correctness of the solution process and update the hypergraph. Additionally, we propose PAC to clarify the interaction between the neural and symbolic components.

3. By combining FGeo-Parser, the rule-based solution generator, and FGeo-HyperGNet, we develop a geometric problem solving system that takes raw geometric problem images and text as input, generates a human-like solution as output, and ensures the complete correctness of the solution.

4. We conducted extensive experiments on the FormalGeo7K \cite{zhang2024formal} dataset, achieving a theorem prediction accuracy (TPA) of 93.50\% and a problem-solving success rate (PSSR) of 88.36\%. Furthermore, we performed ablation studies on the training methods and model architecture of HyperGNet.

\section{Preliminaries}

This section outlines the definition of the problem and models the problem-solving process.

\subsection{Problem Definition and Modeling}

We formulate geometric problems as a collection of conditions and a problem goal and formulate the problem solving process as the sequential application of theorems. Consequently, the process of solving geometric problems can be represented as a hypergraph, where conditions are modeled as hypernodes and theorems as hyperedges. The fundamental terms and examples are defined below.\\
\textbf{Definition 1 Condition ($C$)}: Conditions represent a set of geometric entities, attributes, and relationships. These conditions encompass geometric and quantitative relationships, such as "RightTriangle(ABC)" and "Equal(LengthOfLine(AB),10)".\\
\textbf{Definition 2 Theorem ($T$)}: Theorems constitute pre-defined prior knowledge. A theorem comprises a set of premise conditions and a set of conclusion conditions, both of which are collections of conditions. For instance, the parallel's transitivity can be expressed as  "Parallel(AB,CD) $\&$ Parallel(CD,EF) $\rightarrow$ Parallel(AB,EF)". The collection of all such theorem definitions forms the Prior Knowledge Base $TKB$.\\
\textbf{Definition 3 Goal ($G$)}: Goal represents the objective of geometric problem solving, which can be considered a special form of condition, such as "Value(MeasureOfAngle(ABC))" or "Relation(Parallel(AB,CD))".\\
\textbf{Definition 4 Hypergraph ($H$)}: The solution hypergraph, defined as $H=(C,T,G)$, is a directed hypergraph with known conditions as hypernodes and applied theorems as hyperedges. It describes the structured process of geometric problem solving. A successful application of a theorem can add several new hypernodes to the hypergraph and construct a new hyperedge from a set of premise to a set of new conclusion.

The key for GPS lies in the system's ability to accurately predict the theorem to be applied in the current problem state. Most existing methods \cite{lu2021inter,chen2021geoqa} formulate the theorem prediction task as a generative task, where the parameter optimization objective is to maximize the conditional probability of the next theorem $t_i$ given the previously used theorems $\{t_1, t_2, \dots, t_{i-1}\}$ and the initial hypergraph $h_0$, as shown in Formula~\ref{eq:generative-task}. These methods fail to capture the intermediate states $h_i$ of the problem, thus cannot fully utilize the intermediate results.
\begin{equation}
    \theta^* = \arg\max_{\theta} \prod_{i=1}^{N} P(t_i \mid t_1, t_2, \dots, t_{i-1}, h_0; \theta)
    \label{eq:generative-task}
\end{equation}

Benefiting from the FormalGeo formal system, we can obtain and update the problem state in real time. We formulate the theorem prediction task as a multi-class classification task, where the parameter optimization objective is to maximize the conditional probability of the next theorem $t_i$ given the current problem state $h_{i-1}$, as shown in Formula~\ref{eq:classification-task}. 
\begin{equation}
    \theta^* = \arg\max_{\theta} \sum_{i=1}^{N} P(t_i \mid h_{i-1}; \theta)
    \label{eq:classification-task}
\end{equation}

The geometric problem-solving process can be modeled as a Markov Decision Process \cite{peng2023geodrl}, where the problem solution hypergraph constitutes the state space \( H = \{h_i | i = 0, 1, 2, \dots\} \), and the geometry theorem set constitutes the action space \( T = \{t_i | i = 1, 2, \dots\} \). Given a formal representation of a geometric problem, FormalGeo constructs it into a hypergraph \( h \). The hypergraph \( h_0 \) contains only several unconnected initial condition nodes. Our task is to provide a sequence of theorem \( t \), where each application of \( t_i \) adds new hyperedges and hypernodes to \( h_{i-1} \), thus extends \( h_{i-1} \) to \( h_i \), ultimately constructing a reachable path from the initial conditions to the problem-solving goal. 

\subsection{Predict-Apply Cycle}

\begin{algorithm}
    \caption{Predict-Apply cycle}
    \label{alg:pac}
    \textbf{Input:} $probem$: geometric problems described using formalized language.\\
    \textbf{Output:} $theorem\_seqs$: theorem sequence for problem solving.
    \begin{algorithmic}[1] 
    \STATE Initialize $env$ and $agent$.
    \STATE Initialize $theorem\_seqs$ as $None$.
    \STATE Initialize $applied$ as $True$.
    \STATE $env.init\_hypergraph(problem)$
    \WHILE{$applied$}
        \STATE $hypergraph \leftarrow env.get\_hypergraph()$
        \STATE $theorem \leftarrow agent.predict(hypergraph)$
        \STATE $applied \leftarrow env.apply(theorem)$
        \IF{$env.solved$ is $True$}
            \STATE $theorem\_seqs \leftarrow env.get\_theorem\_seqs()$
            \STATE break
        \ENDIF
    \ENDWHILE
\end{algorithmic}
\end{algorithm}

We have constructed a system comprising a formal environment and a neural agent to accomplish the aforementioned task. This system involves an interaction of two parts, which we refer to as the Predict-Apply Cycle (PAC), as illustrated in Figure~\ref{fig:architecture}. The algorithm is described in Algorithm ~\ref{alg:pac}. The AI agent acquires the current solution hypergraph $h_{i-1}$ of the geometric problem and predicts the theorem $t_{i}$ required for solving the problem. The formalized environment then applies the theorem $t_{i}$, adds new hyperedges and hypernodes, and updates $h_{i-1}$ to $h_{i}$. This interactive process is repeated continuously until the problem is solved or the hypergraph ceases to update. 

\section{Neural-Symbolic Solver}

This section introduces our proposed neural-symbolic architecture, which includes a symbolic formal system built on FormalGeo and a hypergraph neural network based on attention mechanisms.

\subsection{Symbolic System}

Most Existing work has failed to establish a consistent, traceable, and extensible formal system. We have developed a geometric symbolic formal system based on FormalGeo \cite{zhang2024formal}. FormalGeo employs Geometry Definition Language to define the formal system and uses Condition Declaration Language to declare the topological structure of geometric problems, conditions, and problem-solving goals. It first transforms the problem-solving process into the application of geometric theorems, and subsequently further transforms the application process of these theorems into the execution of Geometric Predicate Logic, thereby enabling traceable relational reasoning and algebraic equation solving.

The conditions of geometric problems are stored as quintuples comprising condition ID, condition type, condition body, premises, and theorem. Based on the premises and theorems, we group and structure these conditions, organizing them into a hypergraph with the condition body as hypernodes and the theorem as hyperedges. A set of premise hypernodes and a set of conclusion hypernodes are connected by a theorem hyperedge, thus forming a hypergraph.

This formal system bridges the gap between humans and computers, ensuring that GPS is both human-readable and mathematically rigorous. A more detail discussion of FormalGeo can be found in \cite{zhang2023formalgeo}.

\subsection{Hypernode Encoding}

\begin{figure}
\centering
\includegraphics[width=0.35\textwidth]{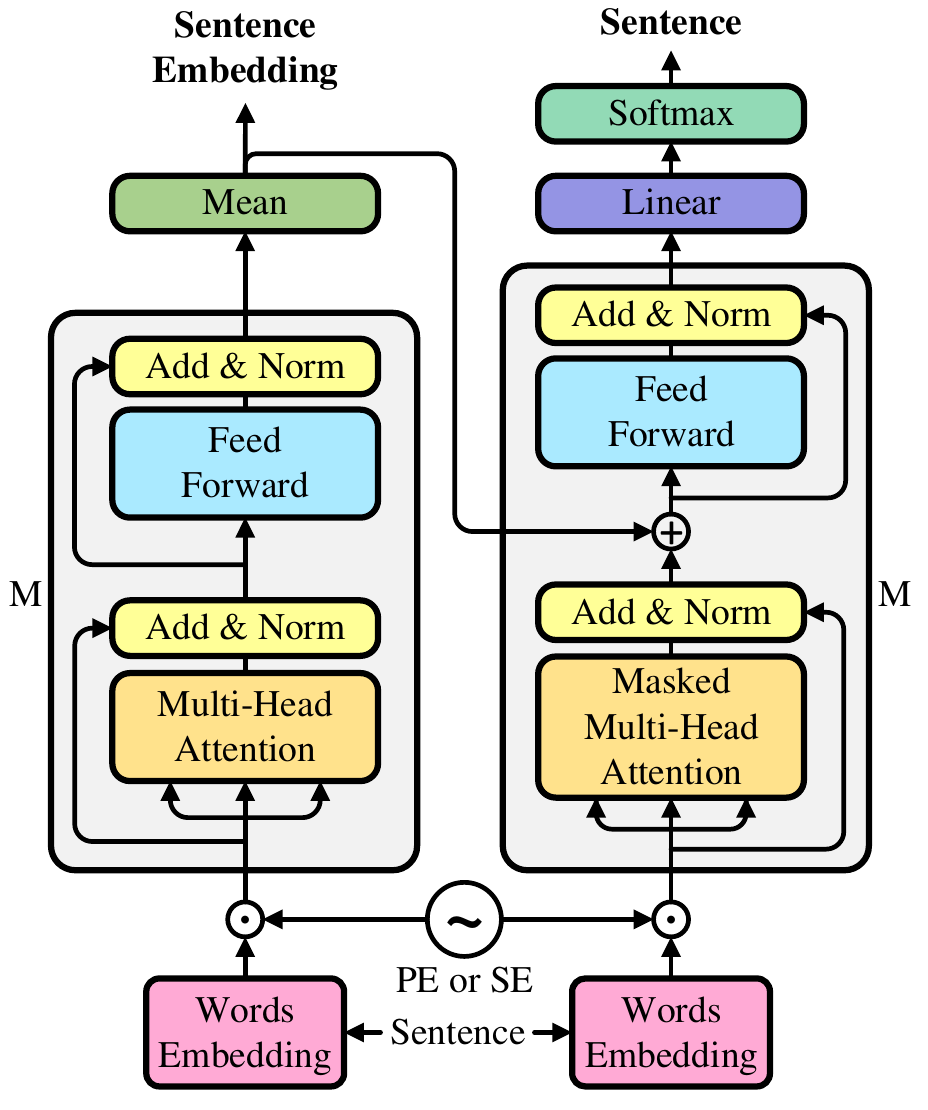}
\caption{The architecture of hypernode encoder. It adopts an encoder-decoder architecture and uses a self-supervised approach to reconstruct the input at hypernode decoder.}
\label{fig:encoder}
\end{figure}

Our task is to create a neural solver that can predict the theorems needed for GPS based on the hypergraph $h_i$ given by the current formal environment. This requires encoding the hypernode and its neighboring hyperedge into a real-valued vector that can be processed by neural networks, which can be viewed as a sentence embedding problem \cite{li2020sentence}. The Transformer \cite{vaswani2017attention} and its derivative network structures are considered powerful architectures for modeling sequential data but are not capable of directly processing graph-structured data. Inspired by Graphormer \cite{ying2021transformers}, We first decompose the semantic and structural information in graph-structured data into a serialized form, and then input and embed it at the appropriate positions in the network.

For a directed hypergraph $h$ containing $n$ hypernodes, we can uniquely represent the hypergraph using the hypernode vector $c = (c_1, c_2, \dots, c_n)$ and the hyperedge theorem adjacency matrix $T_{n\times n}$. $c_i$ represents a condition declaration sentence, composed of a predicate and some individual words, such as "RightTriangle(ABC)". Formal languages and mathematical symbols have domain-specific meanings \cite{ning2023symbolic} and cannot be simply tokenized using natural language tokenization methods. We have designed the FormalGeo tokenizer for neural networks, where \( c_i \) is ultimately represented as a token list, such as \([\text{RightTriangle}, \text{A}, \text{B}, \text{C}]\). The adjacency matrix $T_{n\times n}$ is an extremely sparse matrix, where the element $t_{ij}$ indicates whether there is a hyperedge connecting hypernode $c_i$ and $c_j$. When embedding the row vectors $t_{i}$ of the adjacency matrix $T$, due to its extremely sparse nature, traditional methods would result in the idleness and waste of a large number of neurons. Inspired by the segment encoding of BERT \cite{devlin2019bert}, we remove the empty nodes in \( t_i \) and use position encoding and structure encoding to preserve its structural information. For example, a hyperedge \(t_i = [a, 0, 0, 0, b, 0, 0, 0, c, 0, 0]\) is transformed into \( t_i = [a, b, c] \), \( pe_i = [1, 2, 3] \), and \( se_i = [1, 5, 9] \).

For a hypernode \( c_i \) and its incident hyperedge \( t_i \), we input them into the hypernode encoder, which transforms them into an \( m \)-dimensional vector \( H_i \), as shown in Formula~\ref{eq:encoder}, where HE represents the hypernode encoder and \( \oplus \) represents vector concatenation. We construct the hypernode encoder based on the Transformer architecture, as illustrated in Figure~\ref{fig:encoder}. The hypernode encoder adopts an encoder-decoder architecture and uses a self-supervised approach to reconstruct the input at decoder, thereby forcing the model to learn how to embed semantic and structural information into a fixed-length vector. At the output stage of the encoder, the average of the word sequence embeddings is taken to represent the overall sentence encoding \cite{yan1contrastive}. After passing through the hypernode encoder, the solution hypergraph \( h \) is ultimately encoded into the hypergraph matrix \( H_{n\times m} \) and the goal \( g \in \mathbb{R}^m \).
\begin{equation}
H_i = \text{HE}(c_i + pe_i) \oplus \text{HE}(t_i + pe_i + se_i)
\label{eq:encoder}
\end{equation}

\subsection{HyperGNet Architecture}

\begin{figure}
\centering
\includegraphics[width=0.35\textwidth]{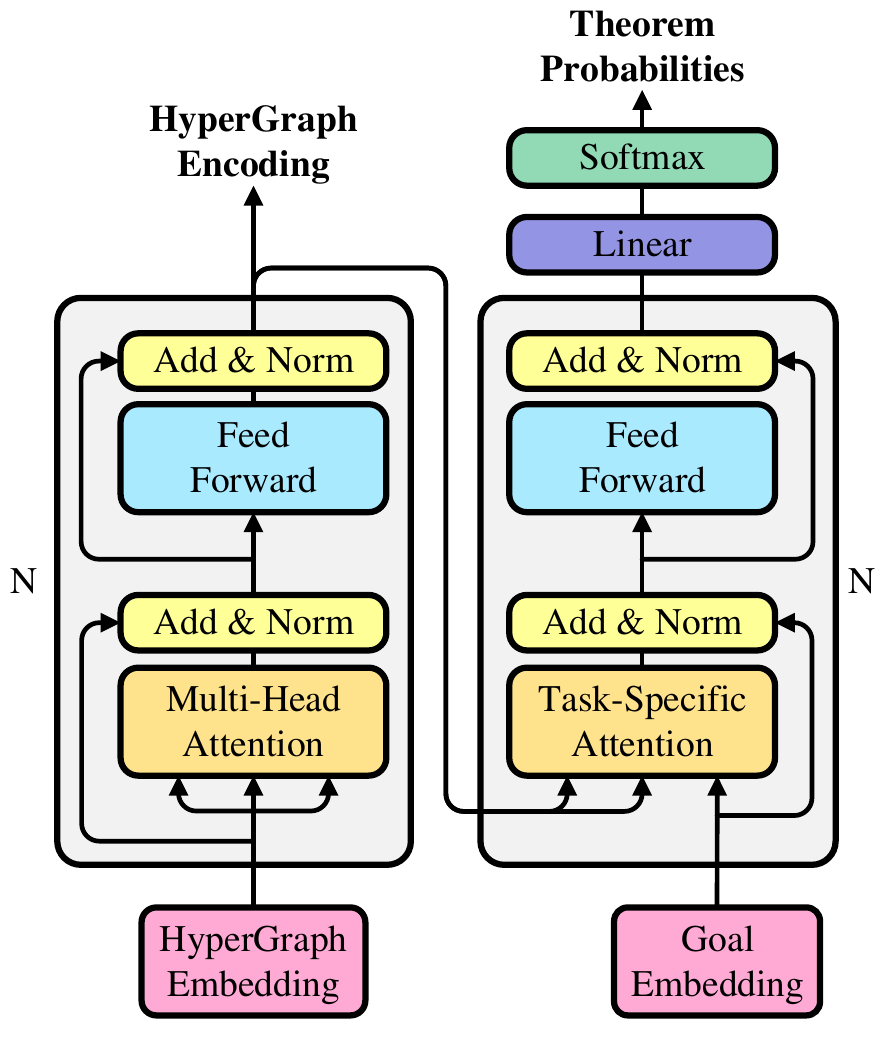}
\caption{The overall architecture of HyperGNet. The left side presents a hypergraph encoder that extracts and fuses hypergraph features. The right side presents a theorem predictor based on task-specific attention.}
\label{fig:predictor}
\end{figure}

As shown in Figure~\ref{fig:predictor}, HyperGNet adopts an encoder-decoder architecture. We use the transformer encoder modules to construct the HyperGNet encoder, and task-specific attention to construct HyperGNet decoder. The solving process of problems can be described with a hypergraph $H=(C,T,G)$, where $C$ and $T$ are processed through hypernode encoder and HyperGNet encoder to obtain the hypergraph encoding $H_{n\times m}^{(N)}$. The goal $G$ can be considered as a special hypernode and embedded into an $m$-dimensional vector $g$.
\begin{equation}
\text{TSA}(Q,K,V) = \text{softmax}\left( Q \times K^T / \sqrt{d_k} \right) \times V
\label{eq:tsa}
\end{equation}

A task-specific attention layer is used to extract key information relevant to solving current problem, as shown in Formula~\ref{eq:tsa}, where $Q = g W^{(Q)}$, $K = H_{n\times m}^{(N)} W^{(K)}$ and $V = H_{n\times m}^{(N)} W^{(V)}$. The other modules of HyperGNet decoder align with Transformer \cite{vaswani2017attention}. In summary, hypergraph $H$ is encoded by the hypernode encoder into hypernode embedding $H_{n\times m}$ and goal embedding $g$, which are fed into HyperGNet to predict the theorem probability $\hat{\vec{y}}$.

For the solution hypergraph $H$, the application process of theorems can be represented as a directed acyclic graph. Multiple alternative theorems may exist at intermediate stages of GPS. We formulate theorem prediction as multi-class classification, decomposing it into binary classification tasks with cross-entropy loss. The loss function is shown in Formula~\ref{eq:tp}, where $\hat{y}$ is the predicted theorem selection probability, $y$ is the ground truth, $\sigma$ is the sigmoid activation function, and $M$ is the number of defined theorems in $TKB$.
\begin{equation}
\mathcal{L} = -\frac{1}{M} \sum_{i=1}^{M} y_i \cdot \log(\sigma(\hat{y}_i)) + (1 - y_i) \cdot \log(1 - \sigma(\hat{y}_i))
\label{eq:tp}
\end{equation}

\section{Experiments}

This section presents the performance of our neural-symbolic architecture on FormalGeo7K \cite{zhang2024formal}, Geometry3K \cite{lu2021inter} and GeoQA \cite{chen2021geoqa}. We compare and analyze the differences in PSSR and TPA between existing methods and the approach proposed in this paper. Additionally, we conduct ablation experiments on the training method and model architecture of HyperGNet.

\begin{table*}
    \centering
    \begin{tabular}{ccccccccc}
        \toprule
        Method & Strategy & Total & $L_1$ & $L_2$ & $L_3$ & $L_4$ & $L_5$ & $L_6$ \\
        \midrule
        Forward Search \cite{zhang2024fgeosss}        & RS  & 39.71 & 58.47 & 41.01 & 34.16 & 16.4  & 5.45  & 4.79 \\
        Backward Search \cite{zhang2024fgeosss}       & BFS & 35.44 & 66.43 & 34.98 & 11.78 & 6.56  & 6.09  & 1.03 \\
        \midrule
        T5-small \cite{raffel2020exploring} with FGeo & BS  & 36.14 & 49.90 & 34.84 & 34.59 & 23.57 & 8.06  & 3.33 \\
        BART-base \cite{lewis2019bart} with FGeo      & BS  & 54.00 & 73.90 & 56.12 & 50.38 & 26.75 & 16.13 & 8.33 \\
        \midrule
        DeepSeek-v3 \cite{deepseekai2024deepseekv3}   & -   & 60.79 & 75.99 & 56.38 & 63.91 & 43.31 & 32.26 & 28.33 \\
        \midrule
        Inter-GPS \cite{lu2021inter}                  & BS  & 60.50 & 76.20 & 63.30 & 60.90 & 39.49 & 17.74 & 15.00 \\
        NGS \cite{chen2021geoqa}                      & BS  & 62.60 & 62.22 & 64.97 & 72.79 & 57.47 & 56.41 & 36.59 \\
        DualGeoSolver \cite{xiao2024learning}         & BS  & 62.11 & 62.96 & 67.80 & 65.44 & 60.92 & 53.85 & 34.15 \\
        FGeo-TP \cite{he2024fgeotp}                   & RS  & 80.86 & 96.43 & 85.44 & 76.12 & 62.26 & 48.88 & 29.55 \\
        FGeo-DRL \cite{zou2024fgeodrl}                & BS  & 80.85 & 97.61 & 91.88 & 70.82 & 57.55 & 36.17 & 27.59 \\
        \midrule
        FGeo-HyperGNet                                & GB  & 88.36 & 96.24 & 91.76 & 87.59 & 82.17 & 56.45 & 56.67 \\
        \bottomrule
    \end{tabular}
    \caption{PSSR of different methods on the FormalGeo7K dataset. Strategy represents different candidate theorem selection methods: BFS stands for Breadth-First Search. DFS stands for Depth-First Search. RS stands for Random Search. BS stands for Beam Search with beam size as \(k\) (\(k=5\)). At each round, the top \(k\) theorems with the highest probabilities are selected, defined as \( \text{TOP}_k \{ p_{i, j} | p_{i, j} = p_i \cdot p_j^{(\text{net}, i)} \}, i \in \{1, 2, \dots, k\}, j \in \{1, 2, \dots, |T|\} \), Where \( p_i \) is the cumulative probability before the beam \( i \), and \( p_j^{(\text{net}, i)} \)is the selection probability of theorem \( j \) predicted by HyperGNet. GB stands for Greedy Beam, which removes any theorems that cannot be applied and adds new applicable theorems to the beam, ensuring that the number of beam heads remains constant. The timeout is set to 600 seconds.}
    \label{tab:detail_results}
\end{table*}

\subsection{Dataset}

We conducted experiments on FormalGeo7K \cite{zhang2024formal}, Geometry3K \cite{lu2021inter} and GeoQA \cite{chen2021geoqa}, partitioning it into a training set, validation set, and test set at a ratio of 3:1:1. We removed the hypernodes from the solution hypergraph, leaving only the hyperedges, which form a directed acyclic graph (DAG) of theorems. Any theorem sequence \( t \) obtained by topological sorting of the theorem DAG can solve the problem. We randomly topological sort the theorem DAG and obtained each step's problem state \( h_{i-1} \) and the set of applicable theorems \( t_i \), yielding data pairs (hypergraph, applicable theorems). This process ultimately generated 20,571 training data pairs (from 4,079 problems), 7,072 validation data pairs (from 1,370 problems), and 7,046 test data pairs (from 1,372 problems).

\subsection{Evaluation Metrics}

We use Theorem Prediction Accuracy (TPA) and Problem-Solving Success Rate (PSSR) as the evaluation metrics to assess different methods. Their definitions are as follows:\\
\textbf{TPA}: Given the current problem state \( h_{i-1} \), the model is tasked with predicting the theorem \( t_i \) required for solving the problem. This metric captures the theorem prediction accuracy of the model at each step of the problem-solving process.\\
\textbf{PSSR}: PSSR is the proportion of successfully solved problems to the total number of problems. Solving a geometric problem requires several theorems, which the model must predict correctly and in the correct order. The predictions are then verified by the formalized system to determine whether the problem is successfully solved. This metric evaluates the model's overall problem-solving capability.

\subsection{Benchmark Methods}

We evaluated a variety of methods on the FormalGeo7K dataset, which include: 1. traditional pure symbolic methods, such as Forward Search and Backward Search; 2. neural-symbolic methods (neural language models integrated with the FormalGeo symbolic system), such as T5-small with FGeo and BART-base with FGeo; 3. pure-neural methods (large language models), including DeepSeek-v3; 4. neural-symbolic systems specifically designed for GPS, such as Inter-GPS, NGS, DualGeoSolver, FGeo-TP, and FGeo-DRL; and 5. our proposed neural-symbolic system, FGeo-HyperGNet.

We also compare the performance of 2 state-of-the-art problem-solving methods on the Geometry3K dataset, GeoDRL \cite{peng2023geodrl} and E-GPS \cite{wu2024egps}; 2 state-of-the-art methods on the GeoQA dataset, SCA-GPS \cite{ning2023symbolic} and DualGeoSolver \cite{xiao2024learning}; 2 state-of-the-art methods on the FormalGeo7K dataset, FGeo-TP \cite{he2024fgeotp} and DFE-GPS \cite{zhang2024diagram}; as well as the performance of FGeo-HyperGNet on the three aforementioned datasets.

The other methods in the comparison use their original parameter settings. For HyperGNet, we first employ a self-supervised approach to pre-train the hypernode encoder. This encoder is then integrated into HyperGNet, allowing for end-to-end training. We set the hidden dimension \( d_{\text{model}} \) of HyperGNet to 256, the number of layers \( N \) to 4, and the number of attention heads \( h \) to 4. Under this configuration, HyperGNet has 20.38 million parameters, significantly fewer than the other methods. During training, we optimize the model parameters using the Adam optimizer, with a learning rate of \( 10^{-5} \), a batch size of 16, and training for 20 epochs. A single training epoch on a GeForce RTX 4090 takes approximately 30 minutes.

\subsection{Experimental Results}

According to the length of the annotated theorem \(l\), we roughly categorize the difficulty of the problems into 6 levels, denoted as \(L_1\)(\(l \leq 2\)), \(L_2\)(\(3 \leq l \leq 4\)), \(L_3\)(\(5 \leq l \leq 6\)), \(L_4\)(\(7 \leq l \leq 8\)), \(L_5\)(\(9 \leq l \leq 10\)), \(L_6\)(\(l \geq 11\)). In Table~\ref{tab:detail_results}, we compare various methods across multiple levels of difficulty on the FormalGeo7K dataset.

Among all evaluated methods, FGeo-HyperGNet achieves the highest overall PSSR of 88.36\% and consistently excels across all difficulty levels. While most methods experience a significant drop in performance as the problem difficulty increases (particularly for \( L_5 \) and \( L_6 \)), FGeo-HyperGNet maintains robust results, achieving 56.67\% on \( L_6 \), compared to 36.59\% by second-best NGS. This highlights its ability to handle challenging geometric problems more effectively than existing approaches.

\setlength{\tabcolsep}{3.5pt}
\begin{table}
    \centering
    \begin{tabular}{cccc}
        \toprule
        Method & Geometry3K & GeoQA & FormalGeo7K \\
        \midrule
        GeoDRL         & 89.40 & -     & -     \\ 
        E-GPS          & 90.40 & -     & -     \\ 
        SCA-GPS        & -     & 64.10 & -     \\ 
        DualGeoSolver  & -     & 65.20 & -     \\ 
        FGeo-TP        & -     & -     & 80.86 \\ 
        DFE-GPS        & -     &  -    & 82.38 \\ 
        FGeo-HyperGNet & 91.99 & 85.64 & 88.36 \\
        \bottomrule
    \end{tabular}
    \caption{PSSR of existing state-of-the-art methods and FGeo-HyperGNet on different datasets.}
    \label{tab:results}
\end{table}
\setlength{\tabcolsep}{6pt}

Traditional search methods achieve limited performance. (Large) Language models perform better but their inability to model geometric structures and relationships limits their ability. FGeo-HyperGNet’s key advantage lies in its integration of neural and symbolic components. AI-driven heuristic search ensures efficient theorem selection, significantly contributing to the method’s superior performance, especially on challenging problems. This design enables it to model long-range dependencies between geometric conditions and theorems effectively. Compared to other neural-symbolic systems, FGeo-HyperGNet significantly outperforming the second-best method, FGeo-TP, by 7.5\%.

We also evaluate the performance of FGeo-HyperGNet against several existing state-of-the-art methods on three datasets: Geometry3K, GeoQA, and FormalGeo7K, as shown in Table~\ref{tab:results}. FGeo-HyperGNet consistently outperforms all competing methods, achieving the highest PSSR across all three datasets. Notably, FGeo-HyperGNet achieves a PSSR of 85.64\%, significantly outperforming the current state-of-the-art (DualGeoSolver with 65.20\%) by 20.44\%, which largely benefits from our proposed formalization approach and neural-symbolic system. FGeo-HyperGNet captures the intermediate states of problems, thereby more accurately modeling the mapping from problems to theorems.

\subsection{Ablation study}

We conduct ablation experiments on the training method and model architecture of HyperGNet, as shown in Table~\ref{tab:ablation_study}. The term \textit{-w/o Pretrain} indicates the removal of the pretraining step, directly proceeding to end-to-end training. \textit{-w/o SE} denotes the exclusion of the structural encoding in Formula~\ref{eq:encoder}. \textit{-w/o Hypergraph} refers to the removal of the hypergraph structural information, where the node sequence is used to represent the problem state. This is implemented by, without altering the network architecture, omitting the hyperedge information and inputting only the node data as sequential information into the network.

\begin{table}
    \centering
    \begin{tabular}{ccccc}
        \toprule
        Method         & Beam Size & TPA & PSSR \\
        \midrule
                        & 1  & 71.58 & 44.86 \\
        FGeo-HyperGNet  & 3  & 88.91 & 62.93 \\
                        & 5  & 93.50 & 67.79 \\
        \midrule
                        & 1  & 70.73 & 41.57 \\
        -w/o Pretrain   & 3  & 87.36 & 59.36 \\
                        & 5  & 92.21 & 64.43 \\
        \midrule
                        & 1  & 70.33 & 39.64 \\
        -w/o SE         & 3  & 88.14 & 60.21 \\
                        & 5  & 92.48 & 64.14 \\
        \midrule
                        & 1  & 68.11 & 36.93  \\
        -w/o Hypergraph & 3  & 87.38 & 57.57 \\
                        & 5  & 92.00 & 63.07  \\
        \bottomrule
    \end{tabular}
    \caption{Ablation study results of HyperGNet on the FormalGeo7k dataset. All ablation experiments used the BS strategy, with a timeout set to 60 seconds.}
    \label{tab:ablation_study}
\end{table}

We observe that as the beam size increases, both TPA and PSSR improve. To assess the effectiveness of pretraining, we removed the pretraining stage and found that TPA and PSSR decreased across all beam size settings. This demonstrates that pretraining is crucial for enabling the hypernode encoder to retain the semantic information of geometric conditions. To evaluate the importance of graph-structured data in the context of GPS, we removed the graph structure information and conducted experiments. As shown in Table~\ref{tab:ablation_study}, we found that both TPA and PSSR decreased, regardless of whether the graph structure was removed alone or if both the graph structure and semantic information were entirely removed (i.e., treating the data as sequential). This indicates that graph-structured data is important for the current task. The information of theorem applications and the relationships between conditions contribute to the improvement of TPA and PSSR.

\subsection{Case Analysis}

\begin{figure}
\centering
\includegraphics[width=0.48 \textwidth]{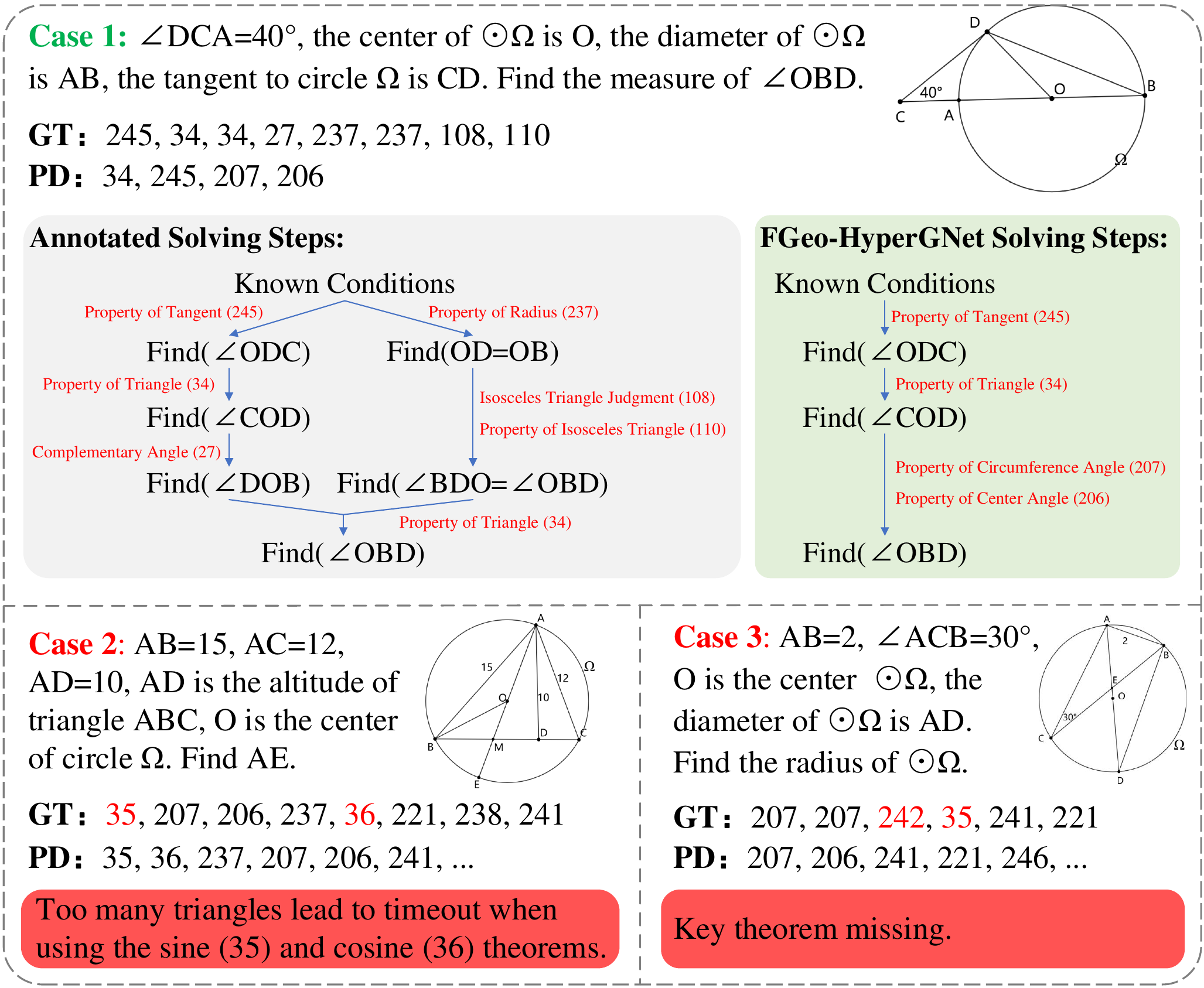}
\caption{Typical cases. Case 1 is a positive case that demonstrates the advantages of FGeo-HyperGNet. Case 2 and 3 are negative cases, providing the reasons why the problem cannot be solved.}
\label{fig:case}
\end{figure}

We select some representative cases in Figure~\ref{fig:case} for further analysis. Taking Case 1 as an example, FGeo-HyperGNet not only provides the solution to the problem but also generates a detailed solution hypergraph. Notably, FGeo-HyperGNet produces a more concise solution process than the human annotations. However, there are some limitations. We found that during the geometric problem-solving process, the time required for solving systems of equations is much longer than that for relational reasoning. The majority of problem-solving failures are due to equation-solving timeouts, especially for theorems involving trigonometric functions, as illustrated in Case 2. Other failures occur because these types of problems are too rare, and the model is unable to adequately learn the solution methods for the minority class from the training data, as shown in Case 3.

\section{Conclusions}

This paper proposes a neural-symbolic architecture for solving formalized plane geometry problems. We are the first to construct a GPS system that takes problem images and text as input, produces a human-like solution, and ensures the complete correctness of the solution. We also achieve state-of-the-art results with a TPA of 93.50\% and a PSSR of 88.36\% on the FormalGeo7K dataset. In the future, we plan to integrate reinforcement learning into the neural component and auxiliary construction into the symbolic component to achieve an automatic IMO-level GPS without human supervision.

\section*{Acknowledgments}
This research was supported by National Natural Science Foundation of China (NSFC) grant 12071282.

\bibliographystyle{named}
\bibliography{ijcai25}

\begin{thebibliography}{}

\bibitem[\protect\citeauthoryear{Buchberger}{1988}]{buchberger1988applications}
Bruno Buchberger.
\newblock Applications of gr{\"o}bner bases in non-linear computational
  geometry.
\newblock {\em Mathematical aspects of scientific software}, pages 59--87,
  1988.

\bibitem[\protect\citeauthoryear{Chen \bgroup \em et al.\egroup
  }{2021}]{chen2021geoqa}
Jiaqi Chen, Jianheng Tang, Jinghui Qin, Xiaodan Liang, Lingbo Liu, Eric Xing,
  and Liang Lin.
\newblock Geoqa: A geometric question answering benchmark towards multimodal
  numerical reasoning.
\newblock In {\em Findings of the Association for Computational Linguistics:
  ACL-IJCNLP 2021}, pages 513--523, 2021.

\bibitem[\protect\citeauthoryear{Chen \bgroup \em et al.\egroup
  }{2022}]{chen2022unigeo}
Jiaqi Chen, Tong Li, Jinghui Qin, Pan Lu, Liang Lin, Chongyu Chen, and Xiaodan
  Liang.
\newblock Unigeo: Unifying geometry logical reasoning via reformulating
  mathematical expression.
\newblock In {\em Proceedings of the 2022 Conference on Empirical Methods in
  Natural Language Processing}, pages 3313--3323. Association for Computational
  Linguistics, 2022.

\bibitem[\protect\citeauthoryear{Chou \bgroup \em et al.\egroup
  }{1995}]{chou1995automated}
Shang-Ching Chou, Xiao-Shan Gao, and Jing-Zhong Zhang.
\newblock Automated production of traditional proofs in solid geometry.
\newblock {\em Journal of Automated Reasoning}, 14(2):257--291, 1995.

\bibitem[\protect\citeauthoryear{Chou \bgroup \em et al.\egroup
  }{2000}]{chou2000deductive}
Shang-Ching Chou, Xiao-Shan Gao, and Jing-Zhong Zhang.
\newblock A deductive database approach to automated geometry theorem proving
  and discovering.
\newblock {\em Journal of Automated Reasoning}, 25(3):219--246, 2000.

\bibitem[\protect\citeauthoryear{DeepSeek-AI}{2024}]{deepseekai2024deepseekv3}
DeepSeek-AI.
\newblock Deepseek-v3 technical report, 2024.

\bibitem[\protect\citeauthoryear{Devlin \bgroup \em et al.\egroup
  }{2019}]{devlin2019bert}
Jacob Devlin, Ming-Wei Chang, Kenton Lee, and Kristina Toutanova.
\newblock {BERT}: Pre-training of deep bidirectional transformers for language
  understanding.
\newblock In Jill Burstein, Christy Doran, and Thamar Solorio, editors, {\em
  Proceedings of the 2019 Conference of the North {A}merican Chapter of the
  Association for Computational Linguistics}, pages 4171--4186, Minneapolis,
  Minnesota, June 2019. Association for Computational Linguistics.

\bibitem[\protect\citeauthoryear{Gan \bgroup \em et al.\egroup
  }{2019}]{gan2019automatically}
Wenbin Gan, Xinguo Yu, Ting Zhang, and Mingshu Wang.
\newblock Automatically proving plane geometry theorems stated by text and
  diagram.
\newblock {\em International Journal of Pattern Recognition and Artificial
  Intelligence}, 33(07):1940003, 2019.

\bibitem[\protect\citeauthoryear{Gelernter}{1959}]{gelernter1959realization}
Herbert~L Gelernter.
\newblock Realization of a geometry theorem proving machine.
\newblock In {\em IFIP congress}, pages 273--281, 1959.

\bibitem[\protect\citeauthoryear{Gowers \bgroup \em et al.\egroup
  }{2023}]{gowers2023artificial}
Timothy Gowers, Po-Shen Loh, Dan Roberts, Geoff Smith, Terence Tao, D.~Sculley,
  Kevin Buzzard, Leo de~Moura, Lester Mackey, and Peter~J. Liu.
\newblock Artificial intelligence mathematical olympiad prize (aimo prize),
  2023.

\bibitem[\protect\citeauthoryear{Guo and Jian}{2022}]{guo2022graph}
Fucheng Guo and Pengpeng Jian.
\newblock A graph convolutional network feature learning framework for
  interpretable geometry problem solving.
\newblock In {\em 2022 International Conference on Intelligent Education and
  Intelligent Research (IEIR)}, pages 59--64. IEEE, 2022.

\bibitem[\protect\citeauthoryear{He \bgroup \em et al.\egroup
  }{2024}]{he2024fgeotp}
Yiming He, Jia Zou, Xiaokai Zhang, Na~Zhu, and Tuo Leng.
\newblock Fgeo-tp: A language model-enhanced solver for euclidean geometry
  problems.
\newblock {\em Symmetry}, 16(4):421, 2024.

\bibitem[\protect\citeauthoryear{Lewis \bgroup \em et al.\egroup
  }{2020}]{lewis2019bart}
Mike Lewis, Yinhan Liu, Naman Goyal, Marjan Ghazvininejad, Abdelrahman Mohamed,
  Omer Levy, Veselin Stoyanov, and Luke Zettlemoyer.
\newblock {BART}: Denoising sequence-to-sequence pre-training for natural
  language generation, translation, and comprehension.
\newblock In Dan Jurafsky, Joyce Chai, Natalie Schluter, and Joel Tetreault,
  editors, {\em Proceedings of the 58th Annual Meeting of the Association for
  Computational Linguistics}, pages 7871--7880, Online, July 2020. Association
  for Computational Linguistics.

\bibitem[\protect\citeauthoryear{Li \bgroup \em et al.\egroup
  }{2020}]{li2020sentence}
Bohan Li, Hao Zhou, Junxian He, Mingxuan Wang, Yiming Yang, and Lei Li.
\newblock On the sentence embeddings from pre-trained language models.
\newblock In {\em Proceedings of the 2020 Conference on Empirical Methods in
  Natural Language Processing (EMNLP)}, pages 9119--9130, 2020.

\bibitem[\protect\citeauthoryear{Littman \bgroup \em et al.\egroup
  }{2022}]{littman2022gathering}
Michael~L Littman, Ifeoma Ajunwa, Guy Berger, Craig Boutilier, Morgan Currie,
  Finale Doshi-Velez, Gillian Hadfield, Michael~C Horowitz, Charles Isbell,
  Hiroaki Kitano, et~al.
\newblock Gathering strength, gathering storms: The one hundred year study on
  artificial intelligence (ai100) 2021 study panel report.
\newblock {\em arXiv preprint arXiv:2210.15767}, 2022.

\bibitem[\protect\citeauthoryear{Lu \bgroup \em et al.\egroup
  }{2021}]{lu2021inter}
Pan Lu, Ran Gong, Shibiao Jiang, Liang Qiu, Siyuan Huang, Xiaodan Liang, and
  Song-chun Zhu.
\newblock Inter-gps: Interpretable geometry problem solving with formal
  language and symbolic reasoning.
\newblock In {\em Proceedings of the 59th Annual Meeting of the Association for
  Computational Linguistics and the 11th International Joint Conference on
  Natural Language Processing (Volume 1: Long Papers)}, pages 6774--6786, 2021.

\bibitem[\protect\citeauthoryear{Nevins}{1975}]{nevins1975plane}
Arthur~J Nevins.
\newblock Plane geometry theorem proving using forward chaining.
\newblock {\em Artificial Intelligence}, 6(1):1--23, 1975.

\bibitem[\protect\citeauthoryear{Ning \bgroup \em et al.\egroup
  }{2023}]{ning2023symbolic}
Maizhen Ning, Qiu-Feng Wang, Kaizhu Huang, and Xiaowei Huang.
\newblock A symbolic characters aware model for solving geometry problems.
\newblock In {\em Proceedings of the 31st ACM International Conference on
  Multimedia}, page 7767–7775. Association for Computing Machinery, 2023.

\bibitem[\protect\citeauthoryear{Peng \bgroup \em et al.\egroup
  }{2023}]{peng2023geodrl}
Shuai Peng, Di~Fu, Yijun Liang, Liangcai Gao, and Zhi Tang.
\newblock Geodrl: A self-learning framework for geometry problem solving using
  reinforcement learning in deductive reasoning.
\newblock In {\em Findings of the Association for Computational Linguistics:
  ACL 2023}, pages 13468--13480, 2023.

\bibitem[\protect\citeauthoryear{Raffel \bgroup \em et al.\egroup
  }{2020}]{raffel2020exploring}
Colin Raffel, Noam Shazeer, Adam Roberts, Katherine Lee, Sharan Narang, Michael
  Matena, Yanqi Zhou, Wei Li, and Peter~J Liu.
\newblock Exploring the limits of transfer learning with a unified text-to-text
  transformer.
\newblock {\em Journal of machine learning research}, 21(140):1--67, 2020.

\bibitem[\protect\citeauthoryear{Sachan and Xing}{2017}]{sachan2017learning}
Mrinmaya Sachan and Eric Xing.
\newblock Learning to solve geometry problems from natural language
  demonstrations in textbooks.
\newblock In {\em Proceedings of the 6th Joint Conference on Lexical and
  Computational Semantics ( SEM 2017)}, pages 251--261, 2017.

\bibitem[\protect\citeauthoryear{Seo \bgroup \em et al.\egroup
  }{2015}]{seo2015solving}
Minjoon Seo, Hannaneh Hajishirzi, Ali Farhadi, Oren Etzioni, and Clint Malcolm.
\newblock Solving geometry problems: Combining text and diagram interpretation.
\newblock In {\em Proceedings of the 2015 conference on empirical methods in
  natural language processing}, pages 1466--1476, 2015.

\bibitem[\protect\citeauthoryear{Trinh \bgroup \em et al.\egroup
  }{2024}]{trinh2024solving}
Trieu~H Trinh, Yuhuai Wu, Quoc~V Le, He~He, and Thang Luong.
\newblock Solving olympiad geometry without human demonstrations.
\newblock {\em Nature}, 625(7995):476--482, 2024.

\bibitem[\protect\citeauthoryear{Tsai \bgroup \em et al.\egroup
  }{2021}]{tsai2021sequence}
Shih-hung Tsai, Chao-Chun Liang, Hsin-Min Wang, and Keh-Yih Su.
\newblock Sequence to general tree: Knowledge-guided geometry word problem
  solving.
\newblock In {\em Proceedings of the 59th Annual Meeting of the Association for
  Computational Linguistics and the 11th International Joint Conference on
  Natural Language Processing (Volume 2: Short Papers)}, pages 964--972, 2021.

\bibitem[\protect\citeauthoryear{Vaswani \bgroup \em et al.\egroup
  }{2017}]{vaswani2017attention}
Ashish Vaswani, Noam Shazeer, Niki Parmar, Jakob Uszkoreit, Llion Jones,
  Aidan~N Gomez, {\L}ukasz Kaiser, and Illia Polosukhin.
\newblock Attention is all you need.
\newblock {\em Advances in neural information processing systems}, 30, 2017.

\bibitem[\protect\citeauthoryear{Wu \bgroup \em et al.\egroup
  }{2024}]{wu2024egps}
Wenjun Wu, Lingling Zhang, Jun Liu, Xi~Tang, Yaxian Wang, Shaowei Wang, and
  Qianying Wang.
\newblock E-gps: Explainable geometry problem solving via top-down solver and
  bottom-up generator.
\newblock In {\em Proceedings of the IEEE/CVF Conference on Computer Vision and
  Pattern Recognition (CVPR)}, pages 13828--13837, June 2024.

\bibitem[\protect\citeauthoryear{Wu}{1978}]{wu1978decision}
W-T Wu.
\newblock On the decision problem and the mechanization of theorem proving in
  elementary geometry.
\newblock {\em Scientia Sinica}, 21:157--179, 1978.

\bibitem[\protect\citeauthoryear{Xiao and Zhang}{2023}]{xiao2023deep}
Ziyang Xiao and Dongxiang Zhang.
\newblock A deep reinforcement learning agent for geometry online tutoring.
\newblock {\em Knowledge and Information Systems}, 65(4):1611--1625, 2023.

\bibitem[\protect\citeauthoryear{Xiao \bgroup \em et al.\egroup
  }{2024}]{xiao2024learning}
Tong Xiao, Jiayu Liu, Zhenya Huang, Jinze Wu, Jing Sha, Shijin Wang, and Enhong
  Chen.
\newblock Learning to solve geometry problems via simulating human
  dual-reasoning process.
\newblock In Kate Larson, editor, {\em Proceedings of the Thirty-Third
  International Joint Conference on Artificial Intelligence, {IJCAI-24}}, pages
  6559--6568. International Joint Conferences on Artificial Intelligence
  Organization, 8 2024.
\newblock Main Track.

\bibitem[\protect\citeauthoryear{Yan \bgroup \em et al.\egroup
  }{2021}]{yan1contrastive}
Y~Yan, R~Li, S~Wang, F~Zhang, W~Wu, and W~ConSERT Xu.
\newblock A contrastive framework for self-supervised sentence representation
  transfer.
\newblock In {\em Proceedings of the 59th Annual Meeting of the Association for
  Computational Linguistics and the 11th International Joint Conference on
  Natural Language Processing}, volume~1, 2021.

\bibitem[\protect\citeauthoryear{Ying \bgroup \em et al.\egroup
  }{2021}]{ying2021transformers}
Chengxuan Ying, Tianle Cai, Shengjie Luo, Shuxin Zheng, Guolin Ke, Di~He,
  Yanming Shen, and Tie-Yan Liu.
\newblock Do transformers really perform badly for graph representation?
\newblock {\em Advances in Neural Information Processing Systems},
  34:28877--28888, 2021.

\bibitem[\protect\citeauthoryear{Zhang \bgroup \em et al.\egroup
  }{1995}]{zhang1995automated}
Jing-Zhong Zhang, Shang-Ching Chou, and Xiao-Shan Gao.
\newblock Automated production of traditional proofs for theorems in euclidean
  geometry i. the hilbert intersection point theorems.
\newblock {\em Annals of Mathematics and Artificial Intelligence},
  13(1-2):109--137, 1995.

\bibitem[\protect\citeauthoryear{Zhang \bgroup \em et al.\egroup
  }{2023a}]{zhang2023lans}
Ming-Liang Zhang, Zhong-Zhi Li, Fei Yin, and Cheng-Lin Liu.
\newblock Lans: A layout-aware neural solver for plane geometry problem.
\newblock {\em arXiv preprint arXiv:2311.16476}, 2023.

\bibitem[\protect\citeauthoryear{Zhang \bgroup \em et al.\egroup
  }{2023b}]{zhang2023multi}
Ming-Liang Zhang, Fei yin, and Cheng-Lin Liu.
\newblock A multi-modal neural geometric solver with textual clauses parsed
  from diagram.
\newblock In {\em Proceedings of the Thirty-Second International Joint
  Conference on Artificial Intelligence, {IJCAI-23}}, pages 3374--3382.
  International Joint Conferences on Artificial Intelligence Organization,
  2023.

\bibitem[\protect\citeauthoryear{Zhang \bgroup \em et al.\egroup
  }{2024a}]{zhang2023formalgeo}
Xiaokai Zhang, Na~Zhu, Yiming He, Jia Zou, Qike Huang, Xiaoxiao Jin, Yanjun
  Guo, Chenyang Mao, Zhe Zhu, Dengfeng Yue, Fangzhen Zhu, Yang Li, Yifan Wang,
  Yiwen Huang, Runan Wang, Cheng Qin, Zhenbing Zeng, Shaorong Xie, Xiangfeng
  Luo, and Tuo Leng.
\newblock Formalgeo: An extensible formalized framework for olympiad geometric
  problem solving, 2024.

\bibitem[\protect\citeauthoryear{Zhang \bgroup \em et al.\egroup
  }{2024b}]{zhang2024fgeosss}
Xiaokai Zhang, Na~Zhu, Yiming He, Jia Zou, Cheng Qin, Yang Li, and Tuo Leng.
\newblock Fgeo-sss: A search-based symbolic solver for human-like automated
  geometric reasoning.
\newblock {\em Symmetry}, 16(4), 2024.

\bibitem[\protect\citeauthoryear{Zhang \bgroup \em et al.\egroup
  }{2024c}]{zhang2024formal}
Xiaokai Zhang, Na~Zhu, Cheng Qin, LI~Yang, Zhenbing Zeng, and Tuo Leng.
\newblock Formal representation and solution of plane geometric problems.
\newblock In {\em The 4th Workshop on Mathematical Reasoning and AI at
  NeurIPS'24}, 2024.

\bibitem[\protect\citeauthoryear{Zhang \bgroup \em et al.\egroup
  }{2024d}]{zhang2024diagram}
Zeren Zhang, Jo-Ku Cheng, Jingyang Deng, Lu~Tian, Jinwen Ma, Ziran Qin, Xiaokai
  Zhang, Na~Zhu, and Tuo Leng.
\newblock Diagram formalization enhanced multi-modal geometry problem solver,
  2024.

\bibitem[\protect\citeauthoryear{Zhu \bgroup \em et al.\egroup
  }{2025}]{zhu2024fgeoparser}
Na~Zhu, Xiaokai Zhang, Qike Huang, Fangzhen Zhu, Zhenbing Zeng, and Tuo Leng.
\newblock Fgeo-parser: Autoformalization and solution of plane geometric
  problems.
\newblock {\em Symmetry}, 17(1), 2025.

\bibitem[\protect\citeauthoryear{Zou \bgroup \em et al.\egroup
  }{2024}]{zou2024fgeodrl}
Jia Zou, Xiaokai Zhang, Yiming He, Na~Zhu, and Tuo Leng.
\newblock Fgeo-drl: Deductive reasoning for geometric problems through deep
  reinforcement learning.
\newblock {\em Symmetry}, 16(4), 2024.

\end{thebibliography}

\end{document}